\documentclass[10pt,twocolumn]{article}

\usepackage[letterpaper,margin=1in]{geometry}

\usepackage[hang]{subfigure}
\usepackage{algorithm,amsmath}
\usepackage{algorithmic}
\usepackage{url}
\usepackage{graphicx}
\usepackage{float}
\usepackage{caption}
\usepackage{amsmath}
\usepackage{amsfonts}
\usepackage[auth-sc,affil-sl]{authblk}

\usepackage{hyperref}
\hypersetup{
	colorlinks,
	breaklinks,
	linkcolor={red!50!black},
	citecolor={blue!50!black},
	urlcolor={blue!80!black},
}

\usepackage[hyperpageref]{backref}
\usepackage{cleveref}
\crefname{section}{Section}{Sections}
\crefname{figure}{Figure}{Figures}
\crefname{table}{Table}{Tables}
\crefname{appendix}{Appendix}{Appendices}
\crefname{definition}{Definition}{Definitions}

\usepackage{scalerel}
\usepackage{tikz}
\usepackage{tikzscale}
\usetikzlibrary{svg.path}

\definecolor{orcidlogocol}{HTML}{A6CE39}
\tikzset{
  orcidlogo/.pic={
    \fill[orcidlogocol] svg{M256,128c0,70.7-57.3,128-128,128C57.3,256,0,198.7,0,128C0,57.3,57.3,0,128,0C198.7,0,256,57.3,256,128z};
    \fill[white] svg{M86.3,186.2H70.9V79.1h15.4v48.4V186.2z}
                 svg{M108.9,79.1h41.6c39.6,0,57,28.3,57,53.6c0,27.5-21.5,53.6-56.8,53.6h-41.8V79.1z M124.3,172.4h24.5c34.9,0,42.9-26.5,42.9-39.7c0-21.5-13.7-39.7-43.7-39.7h-23.7V172.4z}
                 svg{M88.7,56.8c0,5.5-4.5,10.1-10.1,10.1c-5.6,0-10.1-4.6-10.1-10.1c0-5.6,4.5-10.1,10.1-10.1C84.2,46.7,88.7,51.3,88.7,56.8z};
  }
}

\newbox{\myorcidaffilbox}
\sbox{\myorcidaffilbox}{
\resizebox{!}{10pt}{
\begin{tikzpicture}[yscale=-1,transform shape]
\pic{orcidlogo};
\end{tikzpicture}
}
}

\newcommand{\orcidaffil}[1]{%
  \href{https://orcid.org/#1}{\usebox{\myorcidaffilbox}}}

\DeclareMathOperator*{\argmax}{arg\,max}

\begin{document}

\date{}
\title{\textbf{Accelerating Deep Neuroevolution on Distributed FPGAs for Reinforcement Learning Problems}}

\renewcommand\Authands{ and }
\author[]{Alexis Asseman\orcidaffil{0000-0003-4482-5744}}
\author[]{Nicolas Antoine\orcidaffil{0000-0003-0020-1278}}
\author[]{Ahmet S. Ozcan\orcidaffil{0000-0002-4689-7971}}

\affil[]{IBM Almaden Research Center, San Jose, CA, USA.}

\maketitle

\begin{abstract}
  Reinforcement learning augmented by the representational power of deep neural networks, has shown promising results on high-dimensional problems, such as game playing and robotic control.  However, the sequential nature of these problems poses a fundamental challenge for computational efficiency.  Recently, alternative approaches such as evolutionary strategies and deep neuroevolution demonstrated competitive results with faster training time on distributed CPU cores.  Here, we report record training times (running at about 1 million frames per second) for Atari 2600 games using deep neuroevolution implemented on distributed FPGAs.  Combined hardware implementation of the game console, image pre-processing and the neural network in an optimized pipeline, multiplied with the system level parallelism enabled the acceleration.  These results are the first application demonstration on the IBM Neural Computer, which is a custom designed system that consists of 432 Xilinx FPGAs interconnected in a 3D mesh network topology.  In addition to high performance, experiments also showed improvement in accuracy for all games compared to the CPU-implementation of the same algorithm.
\end{abstract}

\section{Introduction}
In reinforcement learning (RL) \cite{arulkumaran2017brief}\cite{li2017deep}, an agent learns an optimal behavior by observing and interacting with the environment, which provides a reward signal back to the agent.  This loop of observing, interacting and receiving rewards, applies to many problems in the real world, especially in control and robotics \cite{polydoros2017survey}. Video games can be easily modeled as learning environments in an RL setting \cite{jaderberg2019human}, where the players act as agents.  The most appealing part of video games for reinforcement learning research is the availability of the game score as a direct reward signal, as well as the low cost of running large amounts of virtual experiments on computers without actual consequences (e.g., crashing a car hundreds of times would not be acceptable).

Deep learning based game playing reached popularity when Deep Q-Network (DQN) \cite{mnih2015human} showed human-level scores for several Atari 2600 games. The most important aspect of this achievement was learning control policies directly from raw pixels in an end-to-end fashion (i.e., pixels to actions). Subsequent innovations in DQN \cite{zhao2016deep}, and new algorithms such as the Asynchronous Advantage Actor-Critic
(A3C) \cite{mnih2016asynchronous} and Rainbow \cite{hessel2018rainbow} made further progress and launched the field to an explosive growth.  A comprehensive and recent review of deep learning for video game playing can be found in \cite{justesen2019deep}. 

However, gradient-based optimization algorithms, used for the training of neural networks, have performance limitations, as they do not lend themselves to parallelization, and they require heavy computations and a large amount of memory, requiring the use of specialized hardware such a Graphical Processing Units (GPU).

Compared to the gradient descent based optimization techniques mentioned above, derivative-free optimization methods such as evolutionary algorithms have recently shown great promise. One of these approaches, called deep neuroevolution, can optimize a neural network's weights as well as its architecture. Recent work in \cite{such2017deep} showed that a simple genetic algorithm with a Gaussian noise mutation can successfully evolve the parameters of a neural network and achieve competitive scores across several Atari games.
Training neural networks with derivative-free methods opens the door for innovations in hardware beyond GPUs. The main implications are related to precision and data flow. Rather than floating point operations, fixed point precision is sufficient \cite{courbariaux2014training} and data flow is only forward (i.e., inference only, no backward flow). Moreover, genetic algorithms are population-based optimization techniques, which greatly benefit from distributed parallel computation.

These observations led us to conclude that genetic algorithm--based optimization of neural networks could be accelerated (and made more efficient) by the use of hardware optimized for fast inference, and the use of multiplicity of such devices would easily take advantage of the inherent parallelism of the algorithm.
Hence, we implemented our solution on the IBM Neural Computer \cite{narayanan2020overview}, which is a custom-designed distributed FPGA system developed by IBM Research. By implementing two instances of the whole application on each of the 416 FPGAs we used (i.e., game console, image pre-processing and the neural net), we were able to run 832 instances in parallel, at an aggregated rate of 1.2 million frames per second.
Our main contributions are:
\begin{itemize}
    \item Introduction of an FPGA-accelerated \emph{Fitness Evaluation Module} consisting of a neural network and Atari 2600 pair, for use with evolutionary algorithms.
    \item The first demonstration of accelerated training quantized neural networks using neuroevolution on distributed FPGAs.
    \item Extensive results on 59 Atari 2600 games trained for six billion frames using deep neuroevolution and performance analysis of our results on the IBM Neural Computer compared to baselines.
\end{itemize}

\section{Related Work}
\label{sec:relatedwork}
Most of the FPGA-based implementations of neural networks target inference applications due to the advantages related to energy efficiency and latency \cite{umuroglu2017finn} \cite{xu2018scaling} \cite{wei2017automated}. These are often based on high-level synthesis for FPGAs, while some of them utilize frameworks that convert and optimize neural network models into bitstreams. FPGA maker Xilinx recently launched a new software platform called Vitis to make it easier for software developers to convert neural network models to FPGA bitstreams.

In addition to the inference-only applications, few studies utilized FPGAs to accelerate reinforcement learning and genetic algorithms. For example \cite{cho2019fa3c} proposed the FA3C (FPGA-based Asynchronous Advantage Actor-Critic) platform which targets both inference and training using single-precision floating point arithmetic in the FPGA.  They show that the performance and energy efficiency of FA3C is better than a high-end GPU-based implementation.  Similar to our work, they chose the Atari 2600 games (only six) to demonstrate their results. However, unlike our work, their Atari 2600 environment is the Arcade Learning Environment \cite{bellemare2013arcade}, which runs on the host CPU.

Genetic algorithms (GA) are another class of optimization methods that FPGA acceleration can help. For example, \cite{tang2004hardware} implemented GA on FPGA hardware and proposed designs for genetic operations, such as mutation, crossover, selection.
Their approach tried to exploit parallelism and pipelining to speed up the algorithm. Experimental results were limited to the optimization of a modified Witte and Holst's Strait Equation, $ f(x_1, x_2, x_3) = |x_1 - a| + |x_2 - b| + |x_3 - c| $, and showed about an order of magnitude speed up compared to a CPU implementation at the time.

A more recent study \cite{torquato2019high} proposed a parallel implementation of GA on FPGAs. They showed results for the optimization of various simple mathematical functions, which are trivial to implement and evaluate in the FPGA itself.  Compared to previous studies, they report speed-up values ranging from one to four orders of magnitude.

Even though these related studies are not a complete picture of the field, our approach is fundamentally different and unique in several aspects.  Rather than accelerating the optimization algorithm (e.g. RL or GA) we have taken a different approach and addressed the data generation (i.e. Atari game environment and obtaining frames). Moreover, we are pipelining the image pre-processing and neural network inference entirely within the FPGA, thus avoiding the costly external memory access, contributing significantly to our results.

\section{Implementation}
\label{sec:implementation}

\subsection{IBM Neural Computer}
\label{sec:inc}

\begin{figure}[t]
  \centering
  \includegraphics[width=\linewidth]{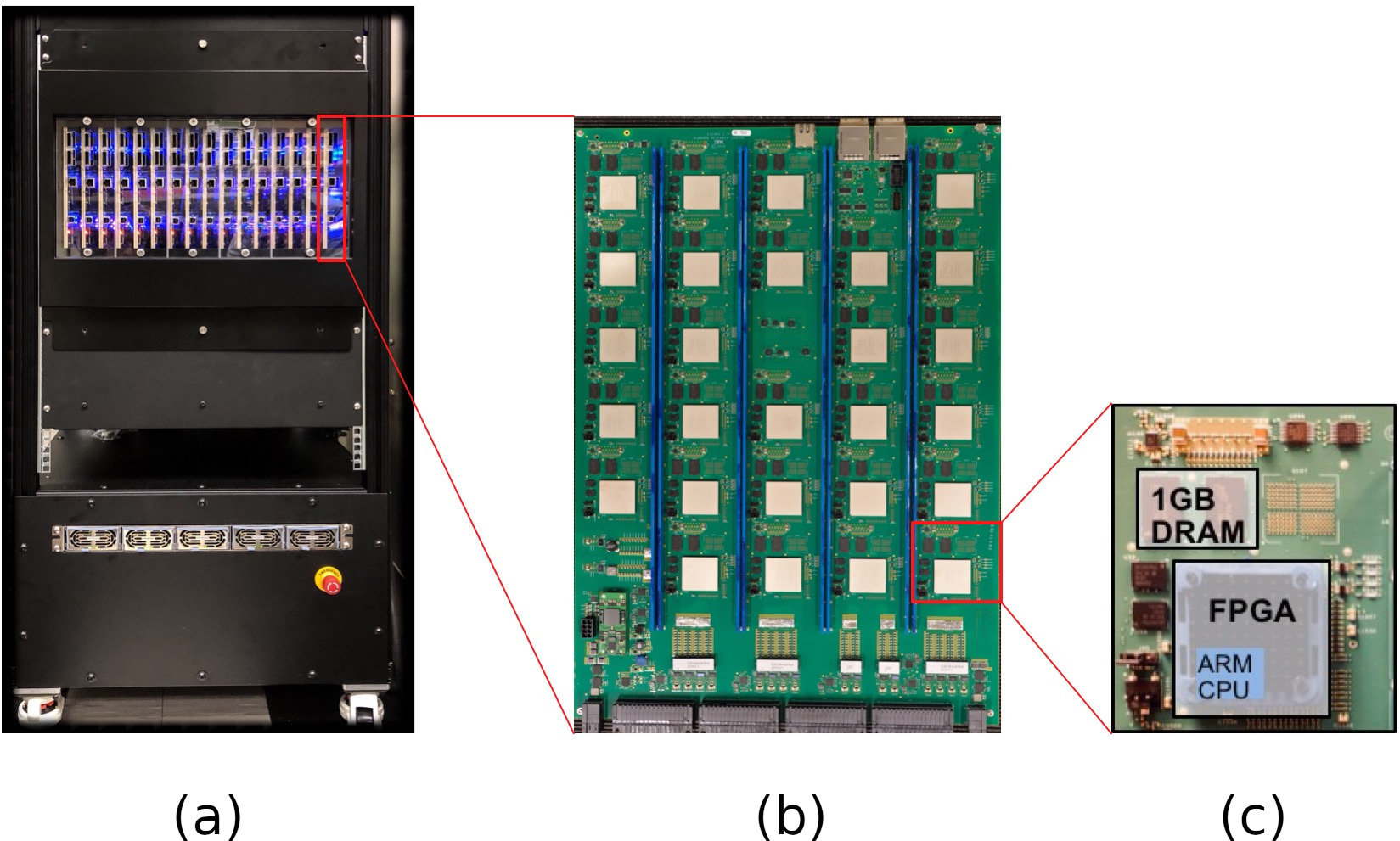}
  \caption{IBM Neural Computer: (a) Cage holding 16 cards (b) Card composed of 27 nodes (c) Node based on a Zynq-7045 with 1GB of dedicated RAM}
  \label{fig:inc}
\end{figure}

The IBM Neural Computer (INC) \cite{narayanan2020overview} is a parallel processing system with a large number of compute nodes organized in a high bandwidth, low latency 3D mesh network. Within each node is a Zynq-7045 system-on-chip, which integrates a dual-core Cortex A9 ARM processor and an FPGA, alongside 1GB of DRAM used both by the ARM CPU and the FPGA.

The INC cage is comprised of a 3D network of $12\times12\times3 = 432$ nodes, which is obtained by connecting 16 cards through a backplane, each containing $3\times3\times3$ nodes (27 nodes per card). The total system consumes about 4kW of power. Each card has one "special" node at coordinate $(xyz)=(000)$ with supplementary control capabilities over its card, and also provides a 4-lane PCIe 2.0 connection to communicate with an external computer.

The 3D mesh network is supported by the high frequency transceivers integrated into the Zynq chip. These are entirely controlled by the FPGA, thus enabling a low level optimization of the network for the target applications.
In particular, the currently implemented network protocols over the hardware network enable us to communicate from any node to any other node of the system, including reading and writing any address accessible over its AXI bus. That last point enables us to control all the Atari 2600 environment fitness evaluation modules present over all the nodes of the system, from the gateway node connected to the computer through PCIe.

We elected to use 26 out of the 27 nodes of each card, leaving the node $(xyz)=(000)$ of each card. Therefore, all the computation carried out in the experiments described herein was on a total of 416 nodes.

\subsection{The Fitness Evaluation Module}

\begin{figure}[t]
  \centering
  \includegraphics[width=\linewidth]{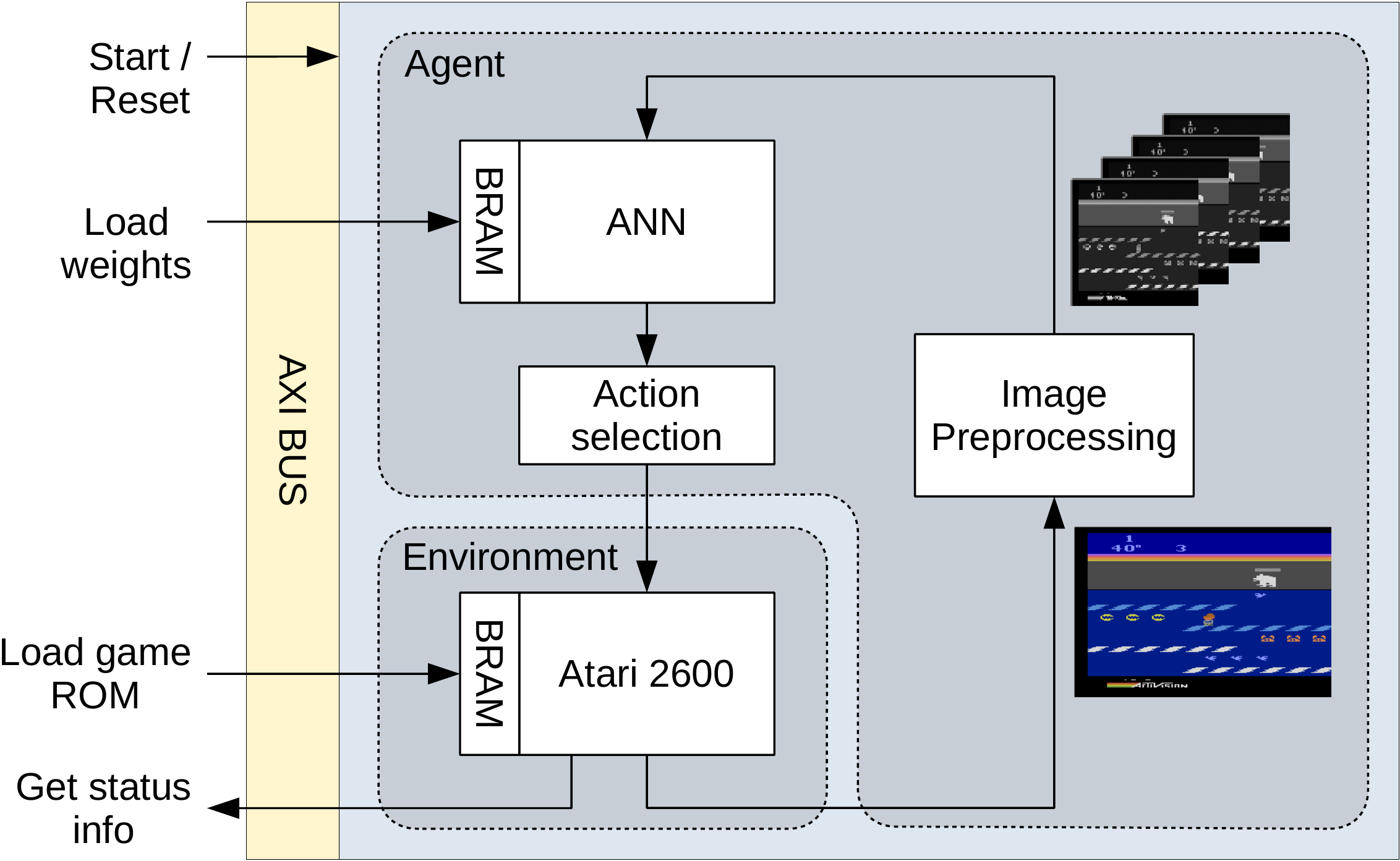}
  \caption{Schematic representation of the fitness evaluation module carrying out the evaluation loop---entirely in FPGA.}
  \label{fig:game_nn_loop}
\end{figure}

\begin{table*}[t]
    \centering
    \caption{Hardware Utilization of a Single Instance of The Fitness Evaluation Module.}
    \begin{tabular}{lrrr}
        \hline
        Submodule                  & Slice LUTs    & BRAM Tiles    & DSPs      \\
        \hline
        Atari 2600              & 1,875         & 9             & 0         \\
        Image pre-processing    & 677           & 16.5          & 2         \\
        Neural network          & 22,855        & 140           & 416       \\
        Miscellaneous           & 1,337         & 0             & 0         \\
        \hline
        Total                   & 26,744        & 165.5         & 418       \\
        \hline
    \end{tabular}
    \label{tab:hw_util}
\end{table*}

The Atari 2600 $\rightarrow$ image pre-processing $\rightarrow$ ANN $\rightarrow$ Atari 2600 loop is integrated in a fitness evaluation module, which can communicate with the AXI bus in order to control the operation from the outside -- i.e. by reading and writing memory-mapped registers exposed on the AXI bus (see fig.~\ref{fig:game_nn_loop}). 

The whole loop is pipelined together, and caching is reduced to the bare minimum to decrease the latency of the loop. Moreover, information exchange between the loop and the rest of the system is done asynchronously, such that the loop is never interrupted by external events. This enabled us to achieve 1,450 frames per second while running the Atari 2600 inside the loop described above.

The module exposes on the AXI address space:
\begin{itemize}
    \item The Atari 2600's block RAM containing the game ROM (write), such that games can be loaded dynamically from the outside.
    \item The ANN's block RAM containing the parameters (write).
    \item The game identifier (write) -- used by the fitness evaluation module to know where in the console's RAM the game status as well as the score are stored.
    \item The status of the game (read) -- Alive or Dead.
    \item The game's score (read).
    \item A frame counter (read).
    \item A clock counter (read) -- to deduce the wall time that passed since the game start.
    \item A command register (write) -- to reset the whole loop (when a new game, new parameters are loaded) and start the game, or to forcibly stop the loop's execution.
\end{itemize}

Table \ref{tab:hw_util} contains a summary of the hardware utilization of the different submodules comprising the fitness evaluation module, as reported by Xilinx's Vivado tool.

We implemented two instances of the fitness evaluation module per INC node, which brings us to a total of 832 instances used in parallel, for a total maximum of 1,206,400 frames per second.

\subsubsection{Atari 2600}

To obtain the highest performance, we chose to avoid software emulation of the Atari 2600 console and took advantage of the FPGA instead, which can easily implement the original hardware functionality of the console at a much higher frequency. We used an open-source VHDL implementation from the open-source MiSTer project\footnote{\url{https://github.com/MiSTer-devel/Main_MiSTer/wiki}}.

We ran the Atari 2600's main clock at 150 MHz, instead of the original 3.58 MHz\cite{stella_guide}. As we are using it in NTSC \cite{pritchard1977us} picture mode, we obtain $\sim2514$ frames per second, compared to 60 frames per second when running the console at its originally intended frequency.  Figure 1 shows snapshots from selected Atari 2600 games.

\begin{figure}[t]
    \center
    \subfigure[Alien]{%
    \includegraphics[width=.3\linewidth]{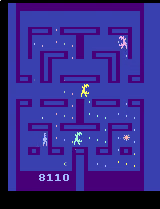}%
    }
    \subfigure[Chopper Command]{%
        \includegraphics[width=.3\linewidth]{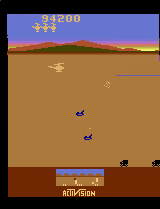}%
    }
    \subfigure[Fishing Derby]{%
        \includegraphics[width=.3\linewidth]{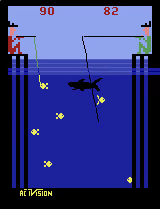}%
    }
    
    \subfigure[Freeway]{%
        \includegraphics[width=.3\linewidth]{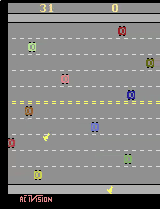}%
    }
    \subfigure[Hero]{%
        \includegraphics[width=.3\linewidth]{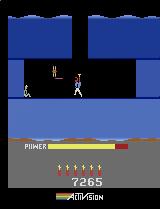}%
    }
    \subfigure[River Raid]{%
        \includegraphics[width=.3\linewidth]{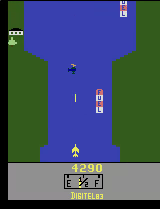}%
    }
    \caption{Screenshots of some of the games we trained on FPGAs using Deep Neuroevolution.}
    \label{fig:game_pictures}
\end{figure}

\subsubsection{Image Pre-processing}

We chose to apply the same image pre-processing as in \cite{mnih2015human} and \cite{such2017deep}, for the dual purpose of enabling an easier comparison with those results, as well as reducing the hardware cost of the ANN (Artificial Neural Network). The entire pre-processing stack is implemented on the FPGA in a pipelined fashion for maximum throughput.
The pre-processing stack is comprised of:
\begin{itemize}
    \item A \textbf{color conversion module} that converts the console's 128 color palette to luminance, using the ITU BT.601 \cite{bt2011studio} conversion standard. This is done instead of just keeping the 3-bit luminance from the console's NTSC signal, such that the 128 color palette is converted to a 124 grayscale color palette (4 levels are lost due to some overlaps in the conversion).
    \item A \textbf{frame-pooling module}. Its purpose is to eliminate sprite flickering (where sprites show on the screen in half of the frames to bypass the sprite limitations of the console). This is achieved by keeping the previous frame in memory, and for each pixel, showing the one that has the highest luminance between the current frame and the previous frame.
    \item A \textbf{re-scaling module}. To re-scale the image from the original $160\times210$ pixels down to $84\times84$ pixels, while applying a bilinear filter to reduce information loss.
    \item A \textbf{frame stacking module}. To stack the frames in groups of 4, where each of the 4 frames becomes a channel of the payload that is fed into the ANN. This has two purposes: It divides the number of inputs to the ANN by 4, and also enables the ANN to see 4 frames at a time, therefore being able to deduce motion within those 4 frames.
\end{itemize}

\subsubsection{ANN Model}

\begin{table*}[t]
    \centering
    \caption{The Artificial Neural Network Architecture for Atari 2600 Action Reward Prediction.}
    \begin{tabular}{lcccccc}
        \hline
        Operation       & Filter size   & Stride        & Output dimensions         & Activation function   & CPF   & KPF   \\
        \hline
        Input image     & -             & -             & $84 \times 84 \times 4$   & -                     & -     & -     \\
        Convolution     & $8 \times 8$  & $4 \times 4$  & $20 \times 20 \times 32$  & ReLU                  & 4     & 32    \\
        Convolution     & $4 \times 4$  & $2 \times 2$  & $9 \times 9\times 64$     & ReLU                  & 32    & 4     \\
        Convolution     & $3 \times 3$  & $1 \times 1$  & $7 \times 7 \times 64$    & ReLU                  & 4     & 32    \\
        Inner product   & -             & -             & $18$                      & -                     & 4     & 1     \\
        \hline
    \end{tabular}
    \label{tab:cnn}
\end{table*}

\begin{table}[t]
    \centering
    \caption{DNNBuilder Fixed-Point Numerical Precision Settings for All Layers.}
    \begin{tabular}{lr}
        \hline
        Bit-width           & 16    \\
        Weights radix       & 13    \\
        Activations radix   & 6     \\
        \hline
    \end{tabular}
    \label{tab:dnnbuilder_params}
\end{table}

The hardware architecture for the neural network was generated using the open-source tool DNNBuilder\footnote{Also known as AccDNN, available at \url{https://github.com/IBM/AccDNN}}\cite{zhang2018dnnbuilder}. It was chosen because it generates human-readable register transfer level (RTL) code, which describes a fully-pipelined neural network, optimized for low block RAM utilization and low latency. DNNBuilder makes this possible by implementing a Channel Parallelism Factor (CPF) and Kernel Parallelism Factor (KPF), which respectively unroll the input and output channels of an ANN layer, at the cost of higher hardware utilization. By alternating the CPF and KPF values at each stage of the ANN, caching, and therefore latency, can be reduced.

Table \ref{tab:cnn} illustrates the architecture of the model, which has been implemented and trained in this study. Note that the model is similar to the one used in \cite{mnih2015human}, but the convolutions are done without padding, and the first fully-connected layer is removed. This was necessary to bring the number of parameters from $\sim 4$ million down to 134,272, such that all the parameters can fit into block RAM for faster access by the ANN modules. Also, we are not using biases since we have not noticed any significant impact on the training performance. Table \ref{tab:dnnbuilder_params} shows the fixed-point numerical precision settings we used for DNNBuilder.

\subsubsection{Action Selection}

The action selection submodule selects the joypad action to apply for the next 4 frames by selecting the action with the maximum reward as predicted by the ANN's output.
To introduce stochasticity into the games, we used \emph{sticky actions} as recommended in \cite{machado2018revisiting}, which introduces stochasticity by having a probability $\varsigma$ of maintaining the action sent to the environment at the previous frame during the current frame, instead of applying the latest selected action. We used the recommended stickiness parameter value $\varsigma = 0.25$. The randomness is sampled from a rather large maximum-length 41-bit linear feedback shift register running independently from the rest of the module.

\subsection{Genetic Algorithm}
\label{sec:geneticalgorithm}

The Genetic Algorithm runs on an external computer, connected to the INC through a PCIe connection that connects it to node (000). 
The node (000) acts as a gateway to the 3D mesh network and enables us to send neural network weights, game ROMs, and start games. It also allows us to gather results from the 832 instances of the fitness evaluation module that are scattered across the 3D mesh network.

The Genetic Algorithm we describe in Algorithm \ref{alg:ga} is largely based upon \cite{such2017deep}. It only includes mutation and selection. Each generation has a population $\mathcal{P}$ that is composed of $N$ individuals. To iterate to the next generation, the top $T$ fittest individuals are selected as parents of the next generation (truncation selection). Each offspring individual is generated from a randomly selected parent with parameters vector $\theta$, to which a vector of random noise is added (mutation) to form the offspring's parameters vector $\theta' = \theta + \sigma \epsilon$, where $\sigma$ is a mutation power hyper-parameter, and $\epsilon$ is a standard normal random vector. Moreover, the fittest parent (elite) is preserved (i.e. unmodified) as individual for the subsequent generation.

\begin{algorithm}[t]
    \caption{Simple Genetic Algorithm}
    \label{alg:ga}
    \begin{algorithmic}
        \STATE {\bfseries Input:} mutation power $\sigma$, population size $N$, number of selected individuals $T$, Xavier random initialization \cite{glorot2010understanding} function $xi$, standard normal random vector generator function $snrv$, fitness function $F$.
        \FOR{$g=1,2...,G \text{ generations}$}
            \FOR{$i=1,...,N-1$ in next generation's population}
                \IF{$g = 1$}
                    \STATE ${\theta}^{g=1}_i = xi()$ \COMMENT{initialize random DNN}
                \ELSE
                    \STATE $k = \text{uniformRandom}(1, T)$ \COMMENT{select parent}
                    \STATE $\theta^{g}_i = \theta^{g-1}_{k} + \sigma * snrv() $ \COMMENT{mutate parent}
                \ENDIF
                \STATE Evaluate $F_i = F(\theta^{g}_i)$
            \ENDFOR
            
            \STATE Sort $\theta^{g}_i$ with descending order by $F_i$
            \IF{$g = 1$}
                \STATE Set Elite Candidates $C \leftarrow \theta^{g=1}_{1...T}$
            \ELSE
                \STATE Set Elite Candidates $C \leftarrow \theta^g_{1...T}\cup \{\text{Elite}\}$
            \ENDIF
                \STATE Set Elite $\leftarrow \argmax_{\theta\in C} \frac{1}{5}\sum_{j=1}^{5}{F(\theta)}$ 
                \STATE $\theta^{g} \leftarrow [\text{Elite}, \theta^{g}-\{\text{Elite}\}]$ \COMMENT{only include elite once}
        \ENDFOR
        \STATE {\bfseries Return: Elite}
    \end{algorithmic}
\end{algorithm}

\section{Experiments}
\label{sec:experiments}

We chose to run the training on 59 out of the 60 games evaluated in \cite{machado2018revisiting}, excluding Wizard Of Wor, which presented some bugs on our Atari 2600 core. The training was carried out in 5 separate experiments to measure the run-to-run variance. Moreover, because the game environment is stochastic, during each run we average the fitness scores of the $T$ fittest individuals over 5 evaluations before selecting the $E$ elites out of those. This procedure helps with generalization of the trained agents. The hyper-parameters of the Genetic Algorithm are presented in table \ref{tab:hyperparams}.

\begin{table}[t]
    \caption{Experimental Hyper-Parameters. Most Were Chosen to Be The Same as in \cite{such2017deep}.}
    \centering
    \begin{tabular}{cc}
        \hline
        Population size ($N$)               & $1000+1$ \\
        Truncation size ($T$)               & $20$ \\
        Number of elites  ($E$)             & $1$\\
        Mutation power ($\sigma$)           & $0.002$ \\
        Survivor re-evaluations             & $5$ \\
        Maximum game time per evaluation    & $5$ minutes \\
        \hline
    \end{tabular}
    \label{tab:hyperparams}
\end{table}

A subset of the results is summarized in Table \ref{tab:table_results_sample}, with the corresponding training plots in Fig.~\ref{fig:training_plots_sample}. 
The complete table of results is available in Appendix \ref{sec:appendix_all_results} in Table \ref{tab:table_results}, along with all the learning plots in Fig.~\ref{fig:training_plots}. 
All of our performance numbers are based on the average and variance over 5 training runs, where each run's performance is based on the average score of the best individual, which was evaluated 5 times. 
We are comparing with DQN (as does \cite{such2017deep}) experiments carried-out in \cite{machado2018revisiting} that use sticky actions as a source of stochasticity as we do. 
We are also comparing with the results from \cite{such2017deep}, which implements very similar experiments in software, with a larger neural network, with the caveat that it uses initial no-ops as a source of stochasticity. 

We are also comparing the approximate wall-clock duration needed to complete a single training experiment with the corresponding algorithms and number of frames. We have measured an evaluation speed of $\sim$ 1 million frames per second, or about 25\% slower than the maximal theoretical speed derived in section \ref{sec:inc}. This is despite running several experiments in parallel to maximize resource utilization and it is largely due to overheads coming from the host computer running the algorithm and communicating with the individual nodes of the INC. Indeed, the current implementation is polling the status of the nodes, and has to send a new set of weights and load the Atari with a new game ROM before starting to evaluate a new individual. This could be further optimized in the future, however, for the current work we chose to avoid the added complexity.

\begin{table*}[t]
    \centering
    \caption{Game Scores for 13 Games From \cite{such2017deep}. The Highest Scores for an Equal Number of Training Frames Are in Bold. Scores Are Averaged Over 5 Independent Training Runs.}
    \begin{tabular}{l|rr|rr|rr}
\hline
                    & DQN \cite{machado2018revisiting}      & GA (ours)         & GA \cite{such2017deep}        & GA (ours)         & GA \cite{such2017deep}    & GA (ours)                     \\
\hline
\# of frames        & \multicolumn{2}{c|}{$200 \cdot 10^6$}                      & \multicolumn{2}{c|}{$1 \cdot 10^9$}              & \multicolumn{2}{c}{$6 \cdot 10^9$}                    \\
Wall clock time     & $\sim$ 10d \cite{such2017deep}     & $\sim$ 6min    & $\sim$ 1h                  & $\sim$ 30min   & $\sim$ 6h              & $\sim$ 2h 30min                \\
\hline
Amidar              & \textbf{792.6}                           & 217.6             & 263                           & \textbf{300.8}       & \textbf{377}                 & 359.8                        \\
Assault             & \textbf{1,424.6}                         & 906.4             & 714                           & \textbf{1,388.2}     & 814                       & \textbf{2,374.6}              \\
Asterix             & \textbf{2,866.8}                         & 1,972.0           & 1,850                         & \textbf{2,616.0}     & 2,255                     & \textbf{2,912.0}             \\
Asteroids           & 528.5                                 & \textbf{2,430.4}     & 1,661                         & \textbf{2,771.6}     & 2,700                     & \textbf{3,227.6}               \\
Atlantis            & \textbf{232,442.9}                       & 55,472.0          & 76,273                        & \textbf{77,832.0}    & 129,167                   & \textbf{136,132.0}                  \\
Enduro              & \textbf{688.2}                           & 76.2              & 60                            & \textbf{100.6}       & 80                        & \textbf{119.6}                   \\
Frostbite           & 279.6                                 & \textbf{3,683.6}     & 4,536                         & \textbf{6,225.2}     & 6,220                     & \textbf{7,241.6}               \\
Gravitar            & 154.9                                 & \textbf{1,056.0}     & 476                           & \textbf{1,636.0}     & 764                       & \textbf{1,948.0}             \\
Kangaroo            & \textbf{12,291.7}                        & 2,564.0           & 3,790                         & \textbf{6,148.0}     & \textbf{11,254}              & 8,232                       \\
Seaquest            & 1,485.7                               & \textbf{2,854.4}     & 798                           & \textbf{3,862.4}     & 850                       & \textbf{5,428}               \\
Skiing              & -12,446.6                             & \textbf{-7,115.2}    & -6,502                        & \textbf{-6,268.6}    & \textbf{-5,541}              & -5,732.6                  \\
Venture             & 3.2                                   & \textbf{908.0}       & 969                           & \textbf{1,052}       & 1,422                     & \textbf{1,428.0}              \\
Zaxxon              & 3,852.1                               & \textbf{5,244.0}     & 6,180                         & \textbf{6,408.0}     & 7,864                     & \textbf{8,324.0}               \\
\hline
\end{tabular}

    \label{tab:table_results_sample}
\end{table*}

\begin{figure*}[t]
\centering
\includegraphics[width=.9\linewidth]{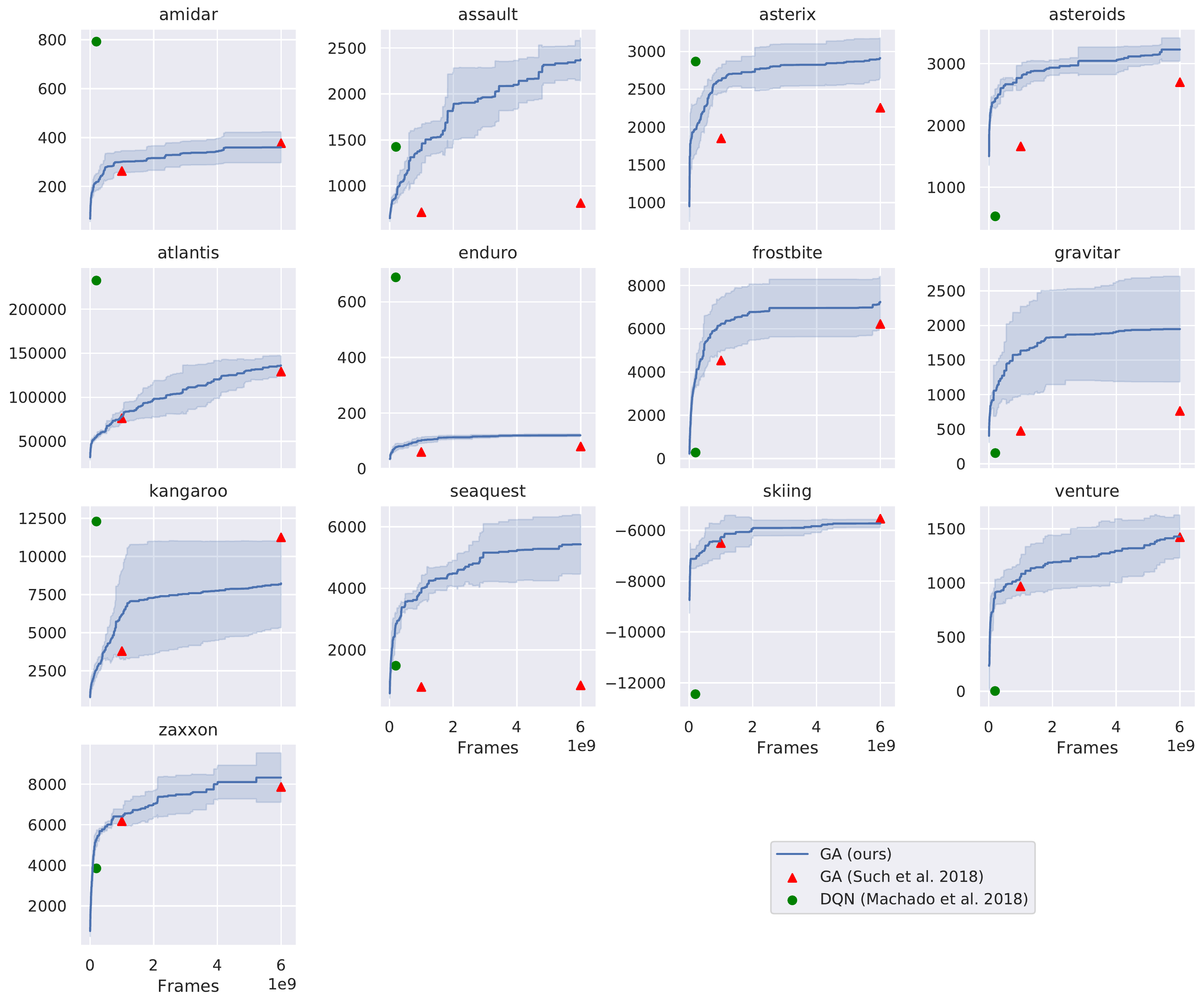}
\caption{GA learning curve across generations on selected games compared to the final scores of DQN at 200M frames (green circles) \cite{machado2018revisiting} and previous GA implementation at 1 and 6B frames (red triangles) from \cite{such2017deep}. Plots for all of the 59 games can be found in the Supplementary section.}
\label{fig:training_plots_sample}
\end{figure*}

\section{Discussion}
The success of a simple GA algorithm in solving complex RL problems was a surprising result \cite{such2017deep} and attracted more research in this area including this work. One of the hypotheses is the improved exploration compared to gradient-based methods.  Potentially GA can avoid being stuck in local minima unlike gradient methods which require additional tricks (e.g., momentum). The promise of GA for training  deep neural networks on reinforcement learning problems also depends on the computational resources.  Even though \cite{such2017deep} showed that the wall clock time can be an order of magnitude smaller compared to RL in learning to play Atari games, the data efficiency does not compare favorably against modern RL methods (e.g. billions of game frames for GA vs. hundreds of millions for algorithms such as A3C). In our work, we attempted to accelerate the game environment and the neural network inference in order to alleviate this bottleneck.

Distributed hardware such as CPUs in the cloud data centers or custom built systems such as ours are a naturally good fit for GA type population-based optimization methods.  Depending on the application, computation vs. communication time needs to be considered carefully.  For example, for game playing, a significant portion of the time is spent during the game itself, which results in a long sequence of inference of game frames and actions.  Communicating game scores and updating neural network weights are sparse in comparison.  Therefore, rather than accelerating the genetic algorithm, acceleration of the game environment and the inference can make a big difference as our results have shown.

The analysis of the game scores agrees with the findings of \cite{such2017deep} and shows that the simple approach of the GA is competitive against a basic RL model such as DQN. Our GA experiments surpass DQN on 30 out of 59 games for an equal number of 200 million training frames, while taking 3 orders of magnitude less wall clock time. When not taking data efficiency into account, GA with 6 billion training frames surpasses DQN with 200 million training frames in 36 out of 59 games, while still taking about 2 orders of magnitude less wall clock time. 

Compared to \cite{such2017deep}, which demonstrated results on thirteen games, we obtained results for 59 games up to six billion frames. Our implementation is about twice as fast as the one in \cite{such2017deep}, which used 720 CPU cores in the cloud. In all instances, our game scores match \cite{such2017deep}, and in some cases even surpass them. Even though the GA algorithm and the experimental hyper-parameters (e.g. population size, mutation power etc.) were identical, the neural network implementations differed. The most significant difference in our implementation is the removal of a fully connected layer and the drastic reduction in the number of weights ($\sim$134k vs. $\sim$4M). One can speculate that the reduced number of parameters was helpful for the GA optimization, however, this needs to be confirmed with an ablation study in the future. Moreover, to introduce stochasticity, we used sticky actions rather than introducing no-ops at the beginning of the game as in \cite{such2017deep}. Indeed, as we have observed experimentally, GA trained models using the random 30 no-op would not generalize to slight perturbations in the game environment, thus invalidating the performance of the trained model. This confirms the findings of \cite{machado2018revisiting} that the random 30 no-op randomization is obsolete, and supports our decision to only present results using sticky actions.

We note that GA failed at hard exploration games such as Montezuma's Revenge or Pitfall. More interestingly, we also note that for games such as Pitfall, Tennis and Double Dunk, the failure was due to the greediness of the algorithm, where initial exploration of the game's mechanics induces a negative score. Therefore the adopted solution is not to act on the game such that the score remains at 0. Pong and Ice Hockey were not affected because the player is not in control of the ball's service.

\section{Conclusion}

In this work, we have shown the acceleration of the fitness evaluation of neural networks playing Atari 2600 games using FPGAs.
Our results were obtained on the recently built IBM Neural Computer, a large distributed FPGA system, demonstrating the advantage of whole application acceleration. 
We used that acceleration with a Genetic Algorithm from \cite{such2017deep} applied to training a deep neural network on Atari 2600 games. 
Compared to the CPU implementation of the neural network in \cite{such2017deep}, the FPGA implementation used a significantly smaller network with quantized weights and activations. The improvements in the game scores compared to \cite{such2017deep} might be due to these differences, which is worth further investigations.
Our results successfully demonstrated that the GA, as a gradient-free optimization method, is an effective way of leveraging the power of hardware that is optimized for limited precision computing and neural network inference. 
We hope to leverage the accelerator to pursue research on gradient-free optimization methods.
Moreover, we are convinced that significant further acceleration and efficiency gains could be achieved with state of the art FPGAs (the Xilinx Zynq-7000 family was released in 2011).

\section*{Acknowledgements}
This paper and the research behind it would not have been possible without the exceptional work and dedication of Chuck Cox (IBM Research) who designed and built the INC system. The authors would also like to thank Winfried Wilcke (IBM Research) for his leadership, support and constant encouragement. Some of the early experiments were run by Miaochen Jin (University of Chicago) during his internship at IBM Research. The authors would like to acknowledge Kamil Rocki (previously at IBM Research) who contributed to the project during its conception.

\bibliographystyle{ieee}
\bibliography{references}

\onecolumn

\appendix
\section{Results on the 59 games}
\label{sec:appendix_all_results}

\begin{table*}[h!]
    \centering
    \caption{Game Scores. All the Scores Are Averaged Over 5 Independent Training Runs. Variance Is Between Parenthesis. The Highest Scores for 200M Frames Are in Bold.}
    \scalebox{0.75}{
        
\begin{tabular}{l|r@{\ \ }lr@{\ \ }l|r@{\ \ }l|r@{\ \ }l}
\hline
                    & \multicolumn{2}{c}{DQN \cite{machado2018revisiting}}      & \multicolumn{2}{c|}{GA}                & \multicolumn{4}{c}{GA}                \\
\hline
\# of frames        & \multicolumn{4}{c|}{$200 \cdot 10^6$}                     & \multicolumn{2}{c|}{$1 \cdot 10^9$}    & \multicolumn{2}{c}{$6 \cdot 10^9$}    \\
Wall clock time     & \multicolumn{2}{c}{$\sim$ 10d \cite{such2017deep}}     & \multicolumn{2}{c|}{$\sim$ 6min}    & \multicolumn{2}{c|}{$\sim$ 30min}   & \multicolumn{2}{c}{$\sim$ 2h 30min}  \\
\hline
Alien               & \textbf{2,742.0}     & (357.5)                               & 1,386.4           & (280.5)           & 1,942.4           & (401.7)           & 3,603.2       & (746.8)        \\
Amidar              & \textbf{792.6}       & (220.4)                               & 217.6             & (34.1)            & 300.8             & (45.0)            & 359.8         & (63.0)        \\
Assault             & \textbf{1,424.6}     & (106.8)                               & 906.4             & (65.6)            & 1,388.2           & (247.9)           & 2,374.6       & (234.4)        \\
Asterix             & \textbf{2,866.8}     & (1,354.6)                             & 1,972.0           & (332.3)           & 2,616.0           & (169.9)           & 2,912.0       & (267.1)        \\
Asteroids           & 528.5             & (37.0)                                & \textbf{2,430.4}     & (157.6)           & 2,771.6           & (197.2)           & 3,227.6       & (187.8)        \\
Atlantis            & \textbf{232,442.9}   & (128,678.4)                           & 55,472.0          & (1,621.4)         & 77,832.0          & (6,786.2)         & 136,132.0     & (10,796.2)        \\
Bank Heist          & \textbf{760.0}       & (82.3)                                & 144.0             & (22.6)            & 205.2             & (39.2)            & 247.2         & (52.1)        \\
Battle Zone         & 20,547.5          & (1,843.0)                             & \textbf{27,000}      & (5,681.5)         & 29,600.0          & (5,128.4)         & 30,680.0      & (5,347.1)        \\
Beam Rider          & \textbf{5,700.5}     & (362.5)                               & 1,276.2           & (122.5)           & 1,442.4           & (215.9)           & 1,486.8       & (266.5)        \\
Berzerk             & 487.2             & (29.9)                                & \textbf{1,020.0}     & (114.7)           & 1,254.8           & (207.3)           & 1,425.6       & (62.2)        \\
Bowling             & 33.6              & (2.7)                                 & \textbf{148.4}       & (18.0)            & 188.2             & (5.8)             & 211.2         & (11.7)        \\
Boxing              & \textbf{72.7}        & (4.9)                                 & 21.6              & (1.5)             & 47.6              & (19.8)            & 70.6          & (13.8)        \\
Breakout            & \textbf{35.1}        & (22.6)                                & 12.8              & (0.4)             & 15.4              & (3.4)             & 18.8          & (5.4)        \\
Carnival            & \textbf{4,803.8}     & (189.0)                               & 4,274.4           & (1,584.6)         & 5,701.2           & (1,581.7)         & 6,268.0       & (1,435.2)        \\
Centipede           & 2,838.9           & (225.3)                               & \textbf{14,629.6}    & (1,710.7)         & 21,163.4          & (2,049.0)         & 25,970.2      & (2,945.4)        \\
Chopper Command     & 4,399.6           & (401.5)                               & \textbf{10,024.0}    & (4,839.8)         & 14,100.0          & (6,566.7)         & 19,932.0      & (9,297.3)        \\
Crazy Climber       & \textbf{78,352.1}    & (1,967.3)                             & 5,896.0           & (1,008.6)         & 11,420.0          & (1,159.0)         & 30,888.0      & (3,243.5)        \\
Defender            & 2,941.3           & (106.2)                               & \textbf{12,216.0}    & (860.8)           & 17,194.0          & (1500.2)          & 20,978.0      & (2,358.2)        \\
Demon Attack        & \textbf{5,182.0}     & (778.0)                               & 2,057.2           & (244.9)           & 2,601.2           & (906.8)           & 3,277.6       & (984.0)       \\
Double Dunk         & -8.7              & (4.5)                                 & \textbf{1.8}         & (0.4)             & 2.0               & (0.0)             & 2.2           & (0.4)        \\
Elevator Action     & 6.0               & (10.4)                                & \textbf{1,764.0}     & (1,131.4)         & 3,360.0           & (1,999.8)         & 6,892.0       & (3,071.3)        \\
Enduro              & \textbf{688.2}       & (32.4)                                & 76.2              & (13.2)            & 100.6             & (9.6)             & 119.6         & (4.3)        \\
Fishing Derby       & \textbf{10.2}        & (1.9)                                 & -49.0             & (6.2)             & -34.2             & (10.1)            & -6.2          & (21.9)        \\
Freeway             & \textbf{33.0}        & (0.3)                                 & 27.4              & (0.5)             & 29.0              & (0.7)             & 29.6          & (1.1)        \\
Frostbite           & 279.6             & (13.9)                                & \textbf{3,683.6}     & (342.2)           & 6,225.2           & (1,226.2)         & 7,241.6       & (1,183.4)        \\
Gopher              & \textbf{3,925.5}     & (521.4)                               & 1,091.2           & (112.4)           & 1,412.0           & (198.9)           & 1,740.0       & (246.6)        \\
Gravitar            & 154.9             & (17.7)                                & \textbf{1,056.0}     & (369.4)           & 1,636.0           & (639.6)           & 1,948.0       & (763.6)        \\
Hero                & \textbf{18,843.3}    & (2,234.9)                             & 10,940.2          & (2,265.1)         & 14,102.8          & (2828.5)          & 17,803.2      & (534.5)        \\
Ice Hockey          & -3.8              & (4.7)                                 & \textbf{10.4}        & (1.3)             & 13.8              & (1.1)             & 15.8          & (1.9)     \\
James Bond          & 581.0             & (21.3)                                & \textbf{1,238.0}     & (479.2)           & 1,778.0           & (454.7)           & 2,670.0       & (569.1)        \\
Journey Escape      & -3,503.0          & (488.5)                               & \textbf{9,556.0}     & (8,335.0)         & 16,980.0          & (8329.7)          & 22,468.0      & (8,340.1)        \\
Kangaroo            & \textbf{12,291.7}    & (1,115.9)                             & 2,564.0           & (506.0)           & 6,148.0           & (2,878.4)         & 8,232         & (2,788.5)        \\
Krull               & \textbf{6,416.0}     & (128.5)                               & 5,875.0           & (1,004.0)         & 7,841.2           & (805.5)           & 10,113.8      & (749.4)        \\
Kung-Fu Master      & 16,472.7          & (2,892.7)                             & \textbf{38,664.0}    & (8,356.2)         & 46,088.0          & (3,588.2)         & 49,616.0      & (2,197.6)        \\
Monzuma's Revenge   & 0.0               & (0.0)                                 & 0.0               & (0.0)             & 0.0               & (0.0)             & 0.0           & (0.0)        \\
Ms. Pacman          & 3,116.2           & (141.2)                               & \textbf{4,004.8}     & (632.1)           & 5,654.4           & (965.4)           & 6,295.6       & (882.9)        \\
Name This Game      & 3,925.2           & (660.2)                               & \textbf{4,388.8}     & (119.7)           & 5,102.8           & (130.2)           & 5,548.4       & (282.8)        \\
Phoenix             & 2,831.0           & (581.0)                               & \textbf{4,846.0}     & (913.5)           & 6,809.6           & (2,096.4)         & 9,957.6       & (2,187.6)        \\
Pitfall             & -21.4             & (3.2)                                 & 0.0               & (0.0)             & 0.0               & (0.0)             & 0.0           & (0.0)        \\
Pong                & \textbf{15.1}        & (1.0)                                 & -16.0             & (2.1)             & -10.4             & (2.5)             & -5.6          & (2.2)        \\
Pooyan              & \textbf{3,700.4}     & (349.5)                               & 1,822.6           & (92.9)            & 2,051.8           & (107.6)           & 2,353.6       & (119.4)        \\
Private Eye         & 3,967.5           & (5,540.6)                             & \textbf{14,996.0}    & (91.7)            & 15,107.0          & (13.3)            & 15,196.6      & (7.6)        \\
Q*bert              & \textbf{9,875.5}     & (1,385.3)                             & 8,378.0           & (3,430.4)         & 9,730.0           & (2,787.8)         & 10,023.0      & (2,438.9)        \\
River Raid          & \textbf{10,210.4}    &  (435.0)                              & 1,919.6           & (722.9)           & 2,642.4           & (874.8)           & 3,502.0       & (674.4)        \\
Road Runner         & \textbf{42,028.3}    &  (1,492.0)                            & 9,744.0           & (939.1)           & 14,848.0          & (2,792.5)         & 21,356.0      & (8,706.1)        \\
Robotank            & \textbf{58.0}        & (6.4)                                 & 20.2              & (1.3)             & 22.4              & (1.9)             & 25.8          & (1.6)        \\
Seaquest            & 1,485.7           & (740.8)                               & \textbf{2,854.4}     & (335.7)           & 3,862.4           & (307.2)           & 5,428         & (966.5)        \\
Skiing              & -12,446.6         & (1,257.9)                             & \textbf{-7,115.2}    & (379.5)           & -6,268.6          & (655.8)           & -5732.6       & (156.9)        \\
Solaris             & 1,210.0           & (148.3)                               & \textbf{4,684.8}     & (964.9)           & 6,201.6           & (1,178.7)         & 8,560.8       & (718.8)        \\
Space Invaders      & 823.6             & (335.0)                               & \textbf{1,175.0}     & (187.9)           & 1,490.6           & (212.6)           & 1,919.8       & (209.3)        \\
Star Gunner         & \textbf{39,269.9}    & (5,298.8)                             & 2,208.0           & (224.8)           & 2,908.0           & (227.0)           & 4,392.0       & (453.6)        \\
Tennis              & -23.9             & (0.0)                                 & 0.0               & (0.0)             & 0.0               & (0.0)             & 0.0           & (0.0)        \\
Time Pilot          & 2,061.8           & (228.8)                               & \textbf{8,388.0}     & (851.7)           & 9,632.0           & (1,227.7)         & 10,620.0      & (1,199.6)        \\
Tutankham           & 60.0              & (12.7)                                & \textbf{157.2}       & (16.7)            & 190.6             & (41.3)            & 213.8         & (49.5)        \\
Up and Down         & 4,750.7           & (1,007.5)                             & \textbf{12,378.4}    & (2,327.8)         & 21,458.8          & (11,005.0)        & 29,244.8      & (14,693.3)        \\
Venture             & 3.2               & (4.7)                                 & \textbf{908.0}       & (125.4)           & 1,052             & (172.4)           & 1,428.0       & (198.3)        \\
Video Pinball       & 15,398.5          & (2,126.1)                             & \textbf{37,039.4}    & (11,059.2)        & 50,880.6          & (13,469.1)        & 62,769.2      & (6,497.0)        \\
Yar's Revenge       & 13,073.4          & (1,961.8)                             & \textbf{26,187.6}    & (3,455.0)         & 34,935.4          & (2,657.7)         & 45,293.2      & (7,313.4)        \\
Zaxxon              & 3,852.1           & (1,120.7)                             & \textbf{5,244.0}     & (359.8)           & 6,408.0           & (366.2)           & 8,324.0       & (1,213.7)        \\
\hline
\end{tabular}

    }
    \label{tab:table_results}
\end{table*}

\begin{figure*}[t!]
    \centering
    \includegraphics[width=\linewidth]{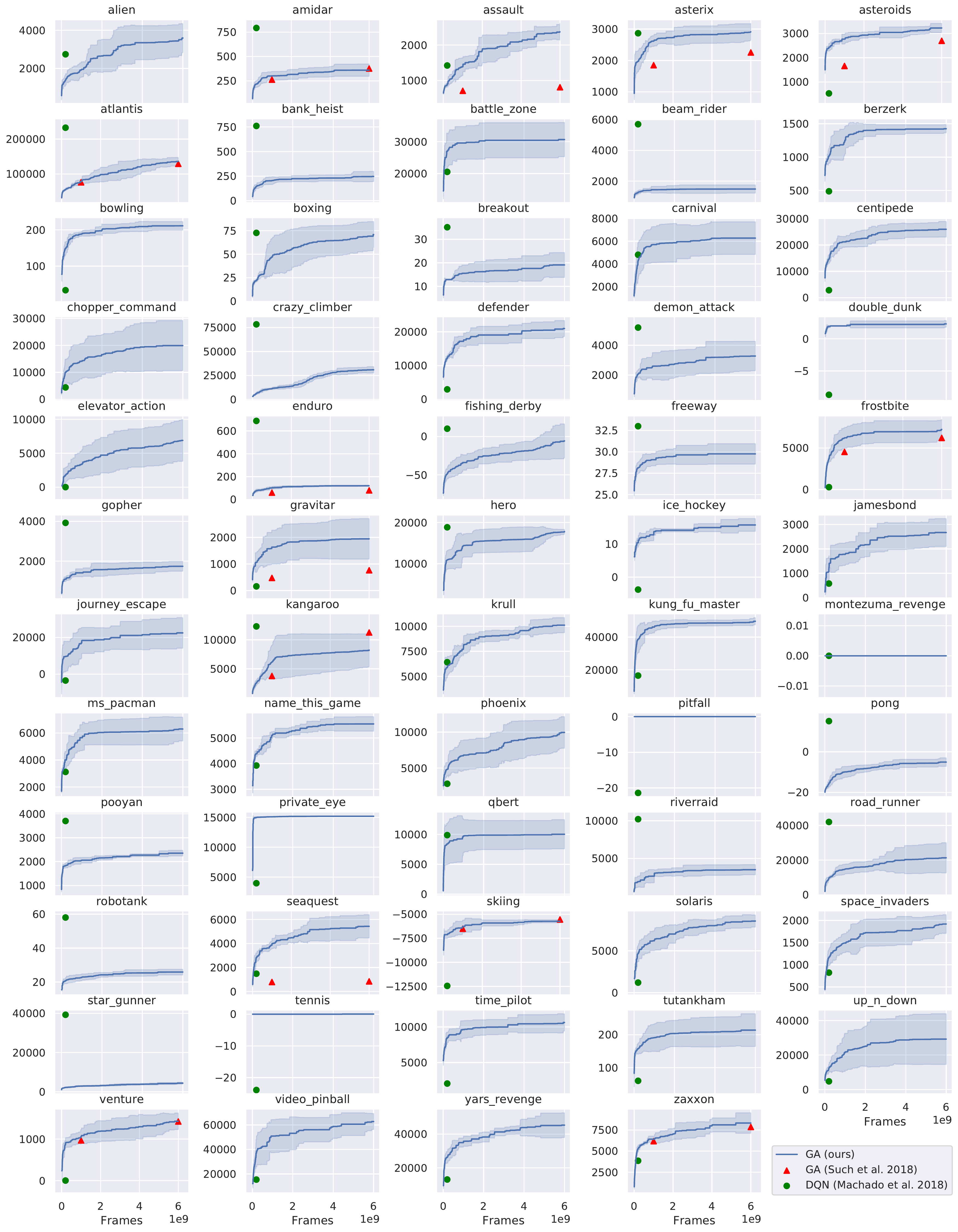}
    \caption{GA learning curve across generations on all games we have run in comparison to the final training scores of DQN at 200M frames \cite{machado2018revisiting} and previous GA implementation at 1B and 6B frames \cite{such2017deep} (only 13 games).}
    \label{fig:training_plots}
\end{figure*}

\end{document}